\documentclass[final]{cvpr}

\usepackage{times}
\usepackage{epsfig}
\usepackage{graphicx}
\usepackage{amsmath}
\usepackage{amssymb}
\usepackage{amsthm}
\usepackage{multirow}

\usepackage[pagebackref=trzue,breaklinks=true,colorlinks,bookmarks=false]{hyperref}

\def\cvprPaperID{5887} 
\def\confYear{CVPR 2021}

\begin{document}

\title{ViPNAS: Efficient Video Pose Estimation via Neural Architecture Search}

\author{Lumin Xu$^{1,2}$ \quad Yingda Guan$^{2}$ \quad Sheng Jin$^{3,2}$ \quad Wentao Liu$^{4}$ \quad Chen Qian$^{4}$ \\ Ping Luo$^{3}$ \quad Wanli Ouyang$^{5}$ \quad Xiaogang Wang$^{1,2}$ \\
$^{1}$ The Chinese University of Hong Kong \quad
$^{2}$ SenseTime Research \quad $^{3}$ The University of Hong Kong \\
$^{4}$ SenseTime Research and Tetras.AI \quad
$^{5}$ The University of Sydney \\
\tt\small luminxu@link.cuhk.edu.hk  \quad \{guanyingda, jinsheng, liuwentao, qianchen\}@sensetime.com \\
\tt\small pluo@cs.hku.hk    \quad wanli.ouyang@sydney.edu.au  \quad xgwang@ee.cuhk.edu.hk
}

\maketitle

\thispagestyle{empty}
\pagestyle{empty}

\begin{abstract}
   Human pose estimation has achieved significant progress in recent years. However, most of the recent methods focus on improving accuracy using complicated models and ignoring real-time efficiency. To achieve a better trade-off between accuracy and efficiency, we propose a novel neural architecture search (NAS) method, termed ViPNAS, to search networks in both spatial and temporal levels for fast online video pose estimation. In the spatial level, we carefully design the search space with five different dimensions including network depth, width, kernel size, group number, and attentions. In the temporal level, we search from a series of temporal feature fusions to optimize the total accuracy and speed across multiple video frames. To the best of our knowledge, we are the first to search for the temporal feature fusion and automatic computation allocation in videos. Extensive experiments demonstrate the effectiveness of our approach on the challenging COCO2017 and PoseTrack2018 datasets. Our discovered model family, S-ViPNAS and T-ViPNAS, achieve significantly higher inference speed (CPU real-time) without sacrificing the accuracy compared to the previous state-of-the-art methods.
\end{abstract}

\section{Introduction}

Human pose estimation has made impressive progress in recent years with the development of stronger neural networks. Most state-of-the-art models~\cite{newell2016stacked,sun2019deep,xiao2018simple} only focus on improving the accuracy, but ignore the computational complexity and real-time performance. However, both accuracy and efficiency are critical for real-world applications of video pose estimation. In this paper, we aim to build a lightweight pose estimator that achieves state-of-the-art performance with significant model complexity reduction. 

\begin{figure}[t]
	\centering
	\includegraphics[width=0.5\textwidth]{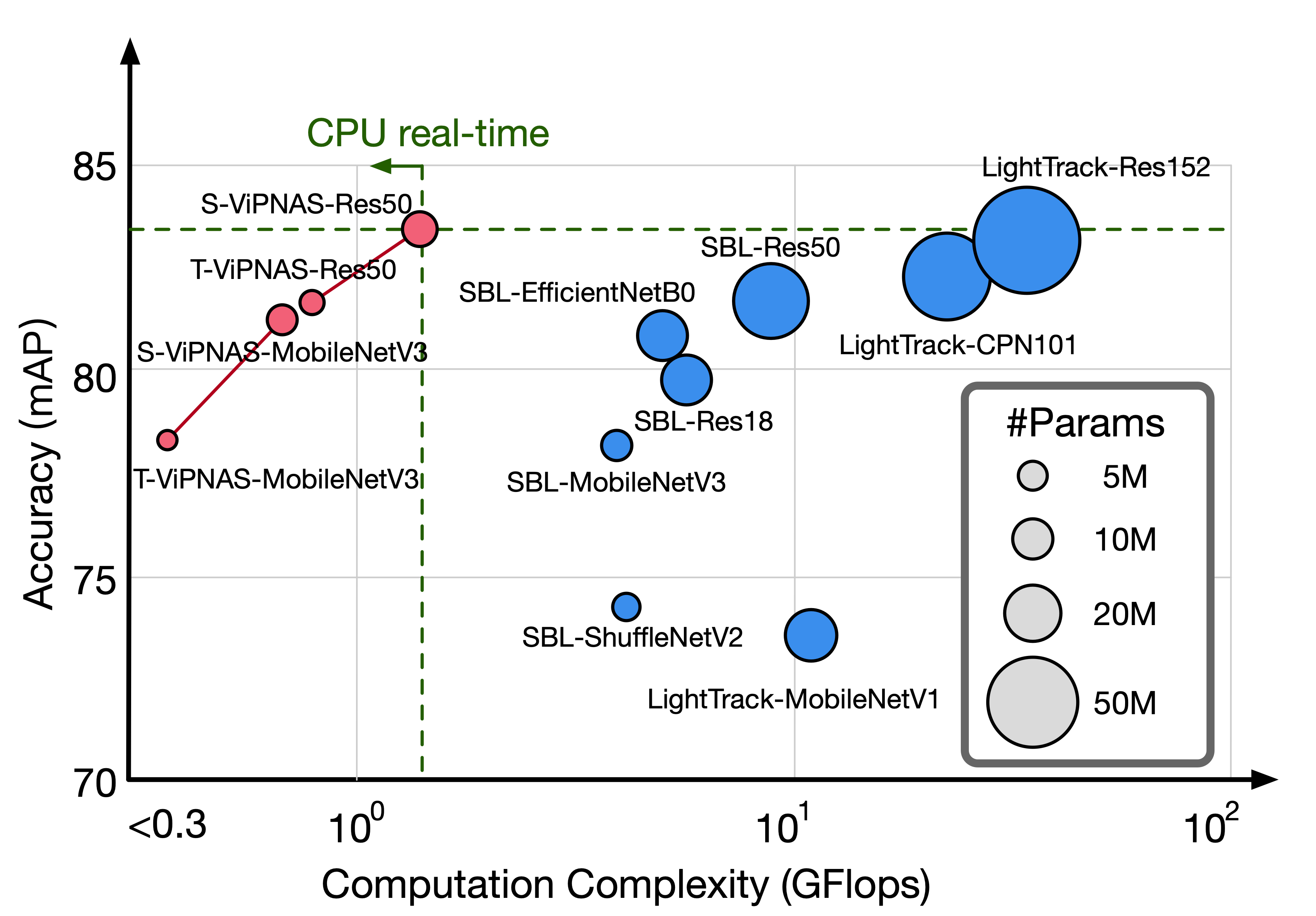}
	\caption{
	\textbf{Speed-accuracy trade-off} on PoseTrack2018~\cite{andriluka2018posetrack} validation set. Methods involve SBL~\cite{xiao2018simple}, LightTrack~\cite{ning2019lighttrack} and our ViPNAS with various backbones. With accuracy comparable to state-of-the-art networks, ViPNAS achieves CPU real-time with significantly lower computation.
	}
	\label{fig:compare}
\end{figure}

For video pose estimation, there is commonly considerable temporal redundancy that leads to superfluous computation, \ie adjacent frames in a video share similar global context information. The temporal contextual information can be used for improving pose estimation. Therefore, it is critical to fuse features from adjacent frames to the current frame in order to effectively utilize the temporal contextual information for balancing accuracy and efficiency. However, there are still several open questions:

1. Low-level local features are important for accurate localization, while higher-level global features are robust to occlusion and large pose variations. Which stage of features should be fused? 

2. For temporal feature fusion, various fusion operations (e.g. addition, multiplication, or concatenation) are chosen by trial-and-error. How to choose the optimal operation?

3. The goal is to optimize the total accuracy subject to the total computation complexity (Flops) constraints over the whole video. Previous works generally explicitly enforce different frames to apply the same model, which will result in sub-optimal performance. How to efficiently allocate computation across different video frames?

Manually exploring the design choices regarding the above questions via trial-and-error can be tedious. We instead apply neural architecture search (NAS) to give a unified solution to them. We propose a novel spatial-temporal NAS framework for efficient video pose estimation, termed ViPNAS. 
For spatial-level search, we optimize the neural architecture by a wide spectrum of five dimensions (depth, width, kernel size, group number, and attention). For temporal-level search, we jointly search three aspects of designs: 1) the stage of features to be fused, 2) the feature fusion operation, and 3) the allocation of computation across video frames. The spatial-level and temporal-level search are jointly optimized through a single framework. Given the total Flops over multiple frames as constraints, we can efficiently allocate computation across different video frames for optimizing performance. Experiments show that ViPNAS significantly improves over the state-of-the-art methods, such as SBL~\cite{xiao2018simple} and LightTrack~\cite{ning2019lighttrack}, with various well-known backbones (ResNet~\cite{he2016deep}, CPN~\cite{chen2018cascaded}, MobileNets~\cite{howard2017mobilenets,howard2019searching}, ShuffleNet~\cite{ma2018shufflenet} and EfficientNet~\cite{tan2019efficientnet}).

Our main contributions can be summarized as follows:
\begin{itemize}
\item We propose the novel spatial-temporal neural architecture search (NAS) framework for efficient video pose estimation, termed ViPNAS. 
\item ViPNAS learns to allocate computational resources (\eg Flops) for different frames under the total computation complexity constraints across frames.
\item ViPNAS automatically searches temporal connections, \ie the fusion module and positions. In the task of video pose estimation, we achieve the state-of-the-art accuracy with CPU real-time performance ($>25$ FPS).
\end{itemize}
\section{Related Work}

\subsection{Human Pose Estimation}

Recent works in human pose estimation~\cite{chen2018cascaded,cheng2020higherhrnet,duan2019trb,jin2020differentiable,jin2020whole,li2019crowdpose,liu2018cascaded,newell2016stacked,sun2019deep,wei2016convolutional,xiao2018simple} focus on designing stronger neural network architectures with higher model capacity to improve accuracy. To better capture the context information, the attention mechanism has been successfully applied in human pose estimation. For example, Chu \etal ~\cite{chu2017multi} proposes multi-context attention to improve model robustness and accuracy. Su \etal ~\cite{su2019multi} proposes SCARB module to enhance pyramid features via spatial and channel-wise context. Other popular attention modules have also been widely explored. For example, Squeeze-and-Excitation (SE) block~\cite{hu2018squeeze} models channel-wise relationship and Global Context (GC) block~\cite{cao2019gcnet} models the global context via addition fusion as NLNet~\cite{wang2018non}. Different from manually design in these works, we propose to apply NAS to automatically search for optimal architectures.

For online video pose estimation, some works~\cite{jin2019multi,jin2017towards,sun2019deep,xiao2018simple,xiu2018pose,xu2020hieve} directly apply the image-based pose models on each video frame. However, such approaches do not capture the temporal consistency, suffering from motion blur or occlusion. Other works utilize temporal cues in order to keep geometric consistency across frames. Such approaches include directly processing concatenated consecutive frames along the channel-axis~\cite{pfister2014deep}, applying 3D temporal convolution~\cite{girdhar2017detect,wang2020combining}, using dense optical flow to produce smooth movement~\cite{pfister2015flowing,song2017thin}. These models are typically computationally expensive, making them not applicable in real-time applications. Recently, some works~\cite{gkioxari2016chained,lihh2019temporal,liwt2019temporal,luo2018lstm,nie2019dynamic} follow the pose propagation paradigm, that transfer features from previous frames to the current frame in an online fashion. However, how to choose the temporal feature fusion sites and the fusion operations are still open questions. We aim to answer this by applying the ViPNAS framework to explore the most effective combination.

\subsection{Neural Architecture Search}

\textbf{NAS for image-level tasks.}
Neural architecture search (NAS) focuses on automating the neural network architecture design. Early NAS approaches~\cite{liu2018progressive,real2019regularized,tan2019mnasnet,zoph2016neural,zoph2018learning} sample a large number of architectures and trained them from scratch, which are very time consuming. Recent NAS approaches~\cite{cai2020once,cai2018proxylessnas,li2020improving,liang2019computation,liu2018darts,liu2020inception,liu2020block,wu2019fbnet,yu2020bignas,zhou2020econas} adopt a weight sharing strategy and train the super-network. Our method also follows this paradigm that trains the super-network only once, and evaluates various sub-networks.

\textbf{NAS for video-level tasks.} 
NAS has been applied in video-level tasks, such as video recognition~\cite{peng2019video,piergiovanni2019tiny,piergiovanni2019evolving,ryoo2019assemblenet}. EVANet~\cite{piergiovanni2019evolving} searches for sequential or parallel model configurations via evolutionary algorithm. AssembleNet~\cite{ryoo2019assemblenet} searches for multi-stream (RGB and optical flow) network connectivity. TinyVideoNet~\cite{piergiovanni2019tiny} searches for computationally efficient classification model for video recognition. 

\textbf{NAS for single-image pose estimation.}
PoseNFS~\cite{yang2019pose} introduces the prior structure of the human body and searches for multiple personalized modules for part-based representations. 
AutoPose~\cite{gong2020autopose} proposes a bi-level optimization method that combines reinforcement learning and gradient-based method.

Our work is different from existing works on NAS in three aspects. First, we are the first to apply NAS for a challenging task of video pose estimation. Second, existing works for image-level and video-level tasks do not search for different architectures at once, but our work search frame-specialized models for further leveraging the ability of NAS in video pose estimation. 
Third, we propose the novel spatial and temporal search space for the task. To fully exploit the temporal information, we search for the optimal combination when fusing features from previous frames to the current frame, which was not explored in these works. Our method also inherits the merit of once-for-all~\cite{cai2020once}, \ie training only once and obtaining many sub-networks, which effectively reduces the searching cost. 

\section{Method}

\subsection{Overview}
\begin{figure*}[t]
	\centering
	\includegraphics[width=0.85\textwidth]{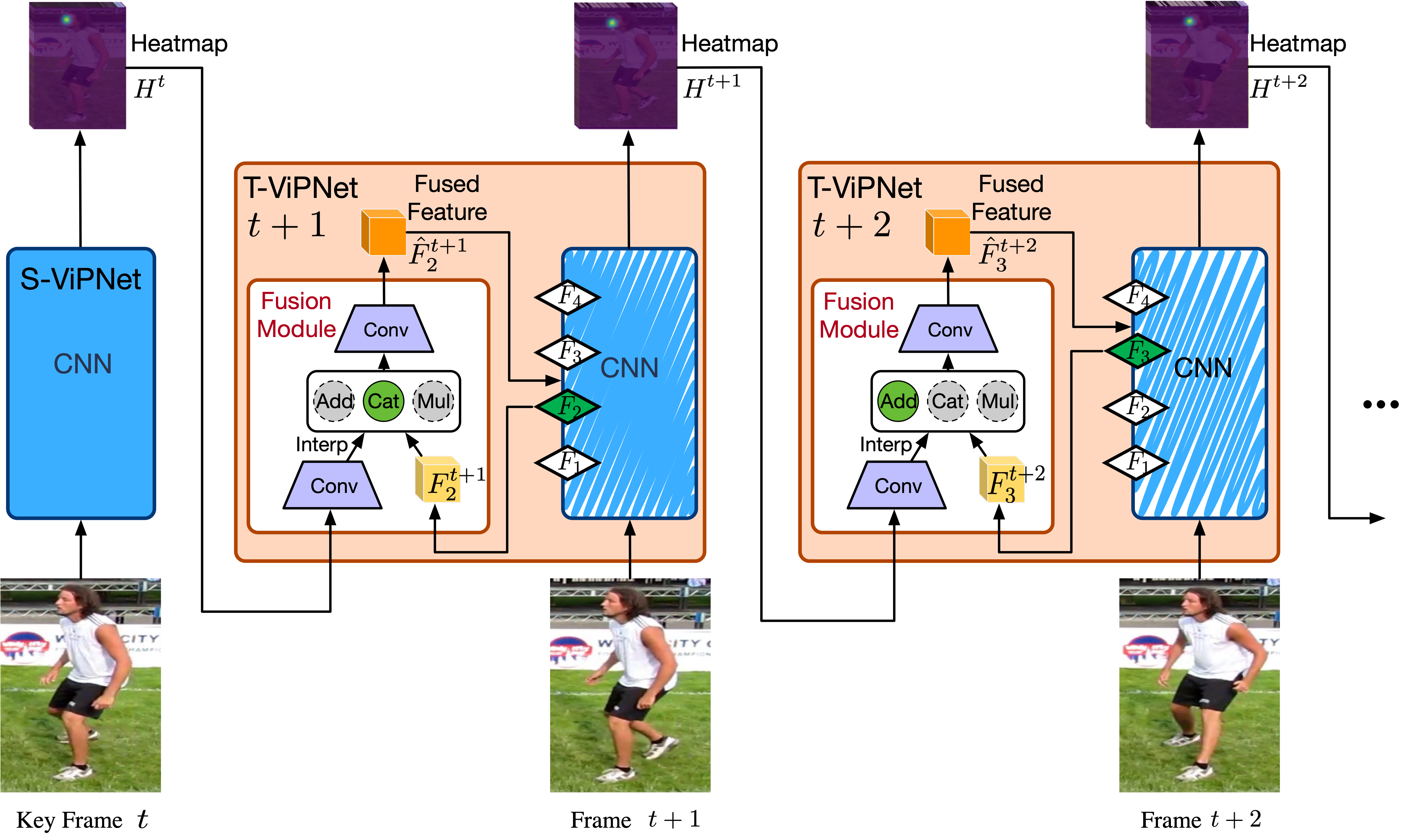}
	\caption{ViPNAS consists of one image-based key frame pose model S-ViPNet, and $T$ video-based pose model T-ViPNets containing temporal feature module and various CNN architectures. Videos are processed frame-by-frame in an online mode. S-ViPNet first predicts the pose heatmaps $H^t$ of the key frame $t$, and propagates them to the next frame $t+1$. T-ViPNet selects the CNN architecture, as well as the input features (\eg $F_1$ to $F_4$) and fusion operation (\eg Add, Cat and Mul) of fusion module. The fusion module combines the selected feature $F_2^{t+1}$ with the propagated heatmaps $H^t$, and generates the fused features $\hat{F}_2^{t+1}$ for predicting the heatmaps $H^{t+1}$.}
	\label{fig:pipeline}
\end{figure*}

Video pose estimation aims to localize the human body parts (also referred to as keypoints or joints) of a person instance in each frame. 

In this paper, based on the online pose propagation paradigm, we propose a novel NAS framework for efficient video pose estimation (ViPNAS). The pipeline of ViPNAS is shown in Figure~\ref{fig:pipeline}. The first frame is selected as the key frame for every $T+1$ frames in the video. For a key frame, a high-accuracy spatial video pose estimation network (S-ViPNet) is applied to localize human poses. We follow the common settings to use heatmaps to encode the joint locations as Gaussian peaks. For a non-key frame, a lightweight temporal video pose estimation network (T-ViPNet) is used for pose propagation. In T-ViPNet, some CNN layers are used for extracting the features of the current frame, then a temporal feature fusion module fuses the features of the current frame and the heatmaps of the last frame. The fused features are then processed by the remaining CNN layers of the T-ViPNet to obtain the heatmaps.  The predicted heatmaps encode the per-pixel likelihood of each joint, which are informative cues to guide the keypoint localization in the subsequent frames. The propagation continues until the next key frame. 

ViPNAS contains two levels of search space, \ie the spatial-level and the temporal-level. The architecture of the key frame (S-ViPNet) is searched in the spatial-level search space. The architectures of the non-key frames (T-ViPNets), including the temporal feature fusion module and CNN layers, are searched in both spatial-level and temporal-level search space. Different non-key frames have different architectures in both the feature fusion module (fusion operation and feature fusion stage) and CNN layers, as shown by the example for frames $t+1$ and $t+2$ in Figure~\ref{fig:pipeline}.

\subsection{Spatial-level Search Space}
 \label{sec:spatial_level}
Motivated by~\cite{cai2020once,yu2020bignas}, we design the weight shared super-network for model architecture search and search for the block number and block structure. Our architecture search spaces extend~\cite{cai2020once,yu2020bignas} to include group and attention for a wider spectrum of five dimensions (depth, width, kernel size, group, and attention). We find out the best configuration of these settings. Our super-network is divided into several stages in series and each stage consists of several blocks having the same spatial resolution of output features. We search on five dimensions as follows:

\textbf{Elastic Depth}: The number of blocks for each stage. We activate the first $D$ blocks of a stage when the depth $D$ is selected for this stage. 
\textbf{Elastic Width}: The number of output channels in each block. We keep the first $W$ filters when the width $W$ is selected.
\textbf{Elastic Kernel Size}: The kernel size of convolutional layers in each block. We reserve the centering $K \times K$ convolutional kernel when the kernel size $K$ is selected. The possible choices of kernel size $K$ are $\{3, 5, 7\}$ for normal convolutional layers and $\{2, 4\}$ for deconvolutional layers.
\textbf{Elastic Group Number}: The group number of convolutional layers~\cite{krizhevsky2017imagenet} in each block. It ranges from $1$ (standard convolution) to $N$ (depth-wise convolution) for $N$ input channels.
\textbf{Elastic Attention Module}:
Using the attention module or not at the end of each block. Since attention modules are shown to be effective for pose estimation in~\cite{chu2017multi,su2019multi},
we include attention modules in our search space. We investigate whether to use the attention module (\eg GC block~\cite{cao2019gcnet} or SE Block~\cite{hu2018squeeze}) at the end of each block. If the attention module is not selected, we skip the attention module and identity mapping is applied. Please refer to Sec.~\ref{sec:supp_spatial_search_space} for more details about the spatial-level search space.

\subsection{Temporal-level Search Space}
\label{sec:temporal_level}

Lightweight pose models alone have difficulty in capturing the global information and distinguishing the joints with similar appearance. However, considering that poses in adjacent video frames are temporally correlated, lightweight models can estimate the joint locations with the local appearance and the guidance from previous frames. 

Temporal feature fusion is critical to the task of video pose estimation, which has also been explored in literature~\cite{gkioxari2016chained,lihh2019temporal,liwt2019temporal,luo2018lstm,nie2019dynamic}. Previous works on temporal fusion mainly differ in two main aspects, \ie the fusion operations and the feature fusion stages. Popular fusion operations may include addition (Add), multiplication (Mul), and concatenation (Cat), \etc. As different pose networks prefer different fusion operations, the choice of the fusion operation is carefully hand-crafted. Besides, different stages of the input features are fused in different approaches. Generally, low-level features may contain more detailed localization information, while higher-level features may contain more global information. In previous works, the levels of features used are mainly chosen by trial-and-error. In ViPNAS, we instead allow the networks to automatically search for the optimal fusion operation and the best stage of features to fuse in a single run of the search. 

As shown in Figure~\ref{fig:pipeline}, our designed temporal feature fusion module includes two inputs, \ie heatmaps of the previous frame and features of the current frame $t+1$. The temporal feature fusion module first selects the location of the input features $F_2^{t+1}$. The pose heatmaps from the adjacent frame are processed by one $1 \times 1$ convolution, followed by one bi-linear interpolation layer to adjust the channels and the resolution (width \& height) to match those of the \emph{selected} features $F_2^{t+1}$. The heatmaps and features are then fused by the \emph{selected} fusion operator (Cat), which are then processed by one $1 \times 1$ convolution, making the fused features ($\hat{F}_2^{t+1}$) have the same shape as the input features. 

Our temporal-level search space for a non-key frame includes $N_O$ choices for the feature fusion operations, \eg addition (Add), multiplication (Mul), and concatenation (Cat); and $N_S$ choices for the input feature stages, \eg $F_1$, $F_2$, $F_3$, and $F_4$. The size of temporal search space is $(N_O \times N_S)^T$, which is impossible to optimize by trial-and-error.

\subsection{Train and Search for ViPNAS}

\subsubsection{Train and Search for S-ViPNet}
Based on the spatial-level search space defined in Section~\ref{sec:spatial_level}, we use the approach in \cite{yu2020bignas} to train the super-network. Sandwich rule~\cite{yu2019universally,yu2020bignas} and in-place distillation~\cite{yu2019universally,yu2020bignas} are applied. Then we sample the sub-networks under the given constraint and evaluate each of them on the validation set to search the architecture of S-ViPNet, which is the network for the key frame. 

\subsubsection{Training for T-ViPNet}
\label{sec:train_tvip}
In this section, we introduce the multi-frame propagation training scheme of our ViPNAS. The goal is to optimize the overall model accuracy at spatial and temporal levels simultaneously in the process of poses propagation across multiple video frames. The overall objective function can be formulated as follows:

\begin{equation}
    \min_{\theta_\mathcal{T}} \sum_{t=1}^T \sum_{\textrm{arch}^t} \mathcal{L}(\mathcal{T}(I^t, H^{t-1} ; \{\theta_\mathcal{T}, \textrm{arch}^t\})),
\label{eq:train_obj}
\end{equation}

\begin{equation}
\textrm{where }  H^{t} = 
\begin{cases}
\mathcal{T}(I^t, H^{t-1} ; \{\theta_\mathcal{T}, \textrm{arch}^t\}),& \quad t\geq 1, \\
\mathcal{S}(I^t ; \{\theta_\mathcal{S}\}), &\quad t = 0.
\end{cases}
\end{equation}
$\mathcal{S}$ is the key frame model S-ViPNet, whose weights are denoted by $\theta_\mathcal{S}$. It is pre-trained and fixed when training and searching for T-ViPNets. $\mathcal{T}$ is the super-network of T-ViPNets, which is parameterized by $\theta_\mathcal{T}$. During training, we sample sub-network consisting of architecture $\textrm{arch}^t$  from $\mathcal{T}$ and the weights of this architecture copied from the super-network weights $\theta_\mathcal{T}$. 
For each frame $t$, $I^t$ is the input image, and $H^t$ is the predicted heatmaps. We use MSE loss function $\mathcal{L}$ to measure the difference between the target heatmaps and the predicted ones of each non-key frame.

\begin{figure}[htb]
	\centering
	\includegraphics[width=0.5\textwidth]{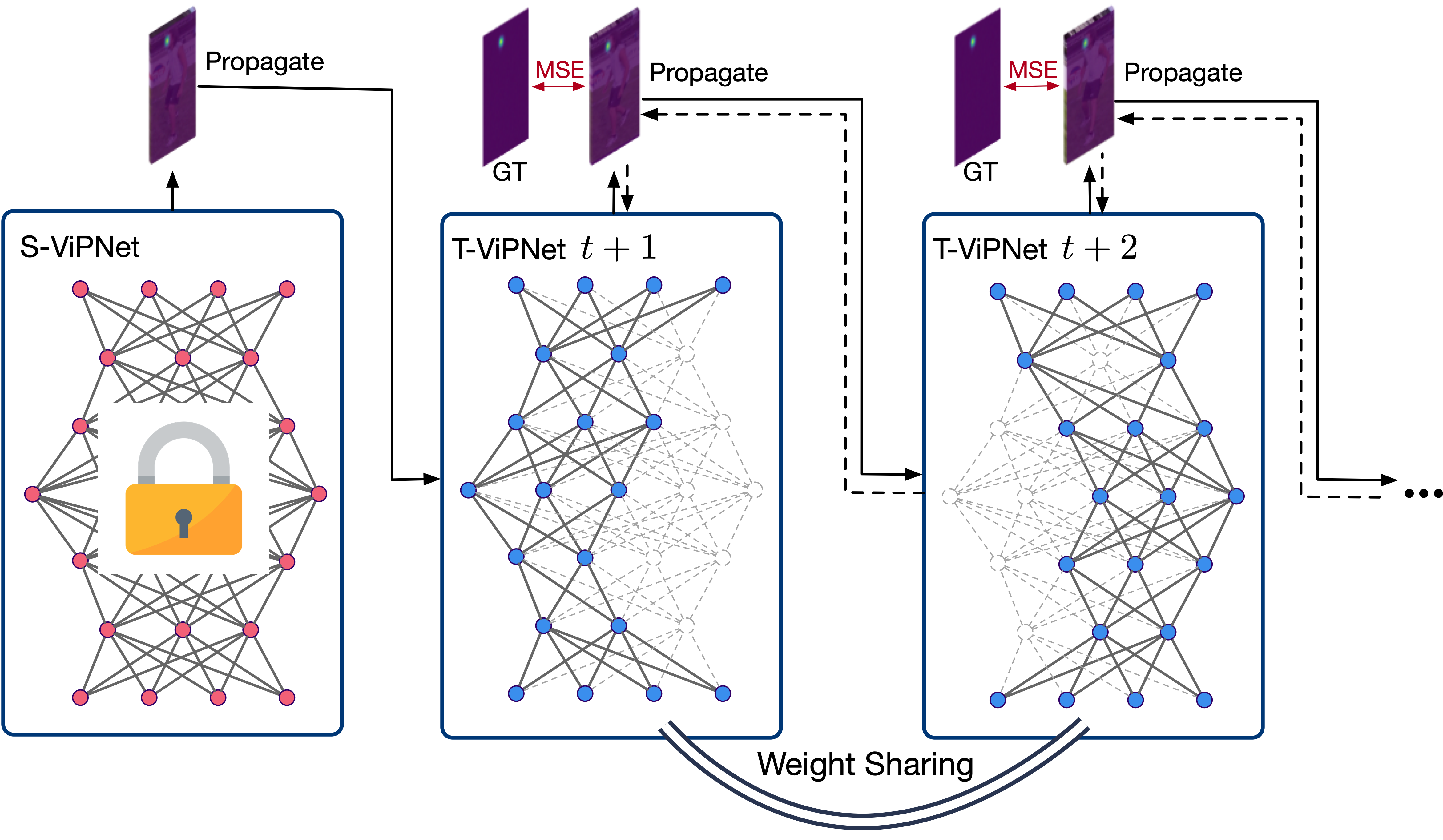}
	\caption{Multi-frame pose propagation training of ViPNAS. The key frame S-ViPNet is first pre-trained and fixed. $T$ ($T \geq 2$) various non-keyframe sub-networks are sampled from a single super-network with sharing weights, and jointly supervised by MSE loss for each frame. The solid lines indicate the forward process and the dotted lines indicate the back propagation process.}
	\label{fig:train&search}
	\vspace{-8pt}
\end{figure}

The multi-frame pose propagation training of T-ViPNet is shown in Figure~\ref{fig:train&search}, where the heatmaps of the key frame are propagated to $T$ ($T \geq 2$) non-key frames iteratively. We apply a single super-network $\mathcal{T}$ for all the non-key frames and all T-ViPNets share the weights, which saves memory during training. Moreover, with one-time training of the super-network, we can search for multiple sets of T-ViPNets with various numbers of propagation frames, see Table.~\ref{tab:propagation_number}. We make the super-network $\mathcal{T}$ to share the same CNN architecture as the discovered S-ViPNet. First, since the tasks of image-based and video-based pose estimation are highly correlated, the good-performing image-based pose estimator can serve as a good candidate architecture for video pose estimation. Second, the pre-trained weights of S-ViPNet can be reloaded for the initialization of the super-network. Third, by sharing similar architectures, the features for the key frame and non-key frames are better aligned.

We jointly train the temporal models (T-ViPNets) in the spatial-level and temporal-level search space for the global optimum. We apply the \emph{Sandwich rule}~\cite{yu2019universally,yu2020bignas} to sample the smallest sub-network, the biggest sub-network and N randomly sampled sub-networks (N = 2 in our experiments) for each mini-batch. We train and search for the CNN architectures and temporal fusion module (including fusion operations and fusion stages) simultaneously. For the biggest (or smallest) sub-network, T-ViPNets of all the frames use the biggest (or smallest) CNN architectures, while the temporal-level search spaces are randomly sampled. For $N$ randomly sampled sub-networks, each T-ViPNet samples unique architecture at both spatial and temporal search spaces. 
\emph{Inplace knowledge distillation}~\cite{yu2019universally,yu2020bignas} takes the prediction of biggest sub-network as the soft labels to enhance supervision for other sub-networks. The biggest sub-network is supervised by ground truth heatmaps with MSE loss, while others are supervised by both the soft labels and the ground truth heatmaps with equal loss weights. 

\vspace{-5pt}
\subsubsection{Automatic Computation Allocation}

As stated above, the sub-networks (T-ViPNets) of different frames do not necessarily share the same architecture. In ViPNAS, different model complexities are assigned to different frames automatically. 

Formally, we aim to search for a group of sub-network architectures ($\{\textrm{arch}^{t}\}_{t=1:T}$) that optimize the overall Average Precision (AP) under the overall computation complexity (Flops) constraints $\textrm{C}$:

\vspace{-5pt}
\begin{equation}
\begin{aligned}
    \max_{\textrm{arch}^{1:T}}& \sum_{t=1}^T  \textrm{AP}(\mathcal{T}(I^t, H^{t-1} ; \{\theta_\mathcal{T}, \textrm{arch}^t\})) \\
\textrm{s.t.} \quad &\sum_{t=1}^T \textrm{Flops}(\textrm{arch}^t) \leq \textrm{C}\\
\end{aligned}
\label{eq:search_obj}
\end{equation}

In the search process, we simply follow~\cite{yu2020bignas} to randomly sample sub-networks that fulfill the given constraints and evaluate the accuracy on the validation set. The sampled sub-networks with the highest AP on the validation set under the Flops constraint are used as the T-ViPNets.

\section{Experiments}

\subsection{Datasets}
\textbf{COCO2017 Dataset}~\cite{lin2014microsoft} is a standard benchmark for human pose estimation. It contains over 200,000 images and 250,000 person instances. We train the models on the COCO train2017 dataset (57K images), and evaluate them on the val2017 set (5K images) and test-dev2017 set (20K images) using the official evaluation metric\footnote{\url{http://cocodataset.org/\#keypoints-eval}}: Average Precision (AP) and Average Recall (AR), which are based on the standard object keypoints similarity (OKS). $\operatorname{OKS} = \frac{\sum_{i}\exp(-d_i^2/2s^2k_i^2)\delta(v_i > 0)}{\sum_i \delta(v_i > 0)}$, where $d_i$ is the Euclidean distance between each ground-truth and the detected keypoint, $v_i$ is the visibility flag, $s$ is the scale of person, and $k_i$ is a constant to control falloff. 

\textbf{PoseTrack2018 Dataset}~\cite{andriluka2018posetrack} is a large-scale dataset for human pose estimation in videos. It contains various videos of human activities with 6 person instances per frame on average. We use PoseTrack2018 V0.25 annotation, which includes 593 training videos, 74 validation videos and 375 testing videos. We follow the common settings~\cite{ning2019lighttrack,sun2019deep,xiao2018simple} to pre-train models on COCO train2017 dataset and fine-tune them on PoseTrack2018 training set. The evaluation follows~\cite{lihh2019temporal,liwt2019temporal,luo2018lstm,nie2019dynamic} for video pose estimation~\cite{jhuang2013towards,zhang2013actemes} that estimates human poses given ground-truth bounding boxes. Pose estimation accuracy is evaluated
using the standard AP metric\footnote{\url{https://posetrack.net/}}.

\subsection{Implementation Details}

We train and search our single-frame pose estimator, termed S-ViPNAS, on COCO dataset. 
For training, we resize the cropped person image to $256 \times 192$, and apply random rotation ([$-40 ^{\circ}$, $40 ^{\circ}$]) and random flip as data augmentation. We train the super-network with inplace knowledge distillation for 250 epochs. Weights are initialized from zero-mean Gaussian distribution with $\sigma$ = 0.001. The basic learning rate is 1e-3, and is reduced by a factor of 10 at the 200th and 230th epoch. We sample 500 models under the Flops constraints and search for S-ViPNAS with the highest AP on the validation set.

We directly transfer the discovered architecture (S-ViPNAS) on the COCO dataset to the PoseTrack dataset. We fine-tune S-ViPNAS on PoseTrack dataset for 20 epochs. The basic learning rate is 1e-4, and drops to 1e-5 at 10 epochs then 1e-6 at 15 epochs. We use S-ViPNAS as the key frame pose estimator and the super-network for temporal propagation models (T-ViPNAS). 
During multi-frame super-network training, the same augmentation methods are applied across $T+1$ frames ($T = 3$ by default). We train the super-network using the sandwich rule for 60 epochs with initial learning rate 1e-3 and cosine learning rate schedule. The search cost is 16 GPU days for training and 2GPU days for search on V100 GPUs.

\subsection{ViPNAS for efficient video pose estimation}
\begin{table*}[tb]
	\begin{center}
	\caption{\textbf{Comparisons} with other video pose estimation approaches on PoseTrack2018 validation set. Our ViPNAS achieves the state-of-the-art performance with significantly lower computation complexity.}
	 \scalebox{0.8}{
		\begin{tabular}{c|c|c|c|c|cccccccc}
			\hline
			Method & Backbone & Image Size & \#Params  &   GFLOPs   &    Head    &    Sho.   &   Elb.   &    Wri.    &    Hip  &    Knee    &    Ank.   &   Total AP   \\ 
			 \hline
			SBL~\cite{xiao2018simple} & ShuffleNetV2~\cite{ma2018shufflenet}  & $256 \times 192$ & 4.7M & 4.21 &84.5 & 83.0 & 74.4 & 63.1 & 74.4 & 70.7 & 63.4 & 63.4 \\
			SBL~\cite{xiao2018simple} & EfficientNetB0~\cite{tan2019efficientnet}  & $256 \times 192$ & 14.9M & 5.05 &  87.3 & 87.9 & 82.0 & 73.4 & 79.0 & 79.2 & 72.9 & 80.7 \\
			SBL~\cite{xiao2018simple} & ResNet-18~\cite{he2016deep}  & $256 \times 192$ & 15.3M & 5.79 & 86.5 & 86.9 & 80.8 & 71.5 & 79.3 & 77.9 & 70.6 & 79.6 \\ \hline
			LightTrack~\cite{ning2020lighttrack} & MobileNetV1~\cite{howard2017mobilenets}  & $384 \times 288$ & 15.8M & 11.3 & 85.2 & 81.7 & 74.7 & 62.9 & 72.9 & 69.4 & 61.4 & 73.4 \\
			LightTrack~\cite{ning2020lighttrack} & CPN101~\cite{chen2018cascaded}  & $384 \times 288$ & 46.3M & 22.9 &  87.9 & 87.7 & 83.5 & 75.8 & 79.1 & 80.4 & 77.0 & 82.1 \\
			LightTrack~\cite{ning2020lighttrack} & ResNet152~\cite{he2016deep}  & $384 \times 288$ & 68.6M & 35.6 & 89.4 & 88.5 & 84.4 & 76.2 & 81.2 & 80.5 & 77.5 & 83.0 \\ \hline
			SBL~\cite{xiao2018simple} & MobileNet-V3~\cite{howard2019searching}  & $256 \times 192$ & 5.5M & 4.1 & 86.4 & 85.9 & 78.8 & 69.6 & 76.5 & 76.1 & 69.1 & 78.1 \\
		    S-ViPNAS & MobileNet-V3~\cite{howard2019searching}  & $256 \times 192$ & 5.4M & 0.69   & 87.8 & 88.0 & 82.3 & 74.1 & 78.8 & 79.1 & 74.0 &  81.1 \\
		    T-ViPNAS & MobileNet-V3~\cite{howard2019searching}  & $256 \times 192$ & 2.5M & \textbf{0.37}   & 87.3 & 85.6 & 78.9 & 70.3 & 75.7 & 75.0 & 70.1 & 78.2 \\
            \hline
			SBL~\cite{xiao2018simple} & ResNet-50~\cite{he2016deep}  & $256 \times 192$ & 34.0M & 8.99  & 86.7 & 88.1 & 83.0 & 75.7 & 80.8 & 80.4 & 74.2 & 81.6 \\ 
			S-ViPNAS & ResNet-50~\cite{he2016deep}  & $256 \times 192$ & 7.3M & 1.44  & 88.1 & 89.6 & 84.5 & 77.4 & 81.1 & 81.8 & 77.6 & \textbf{83.2} \\
            T-ViPNAS & ResNet-50~\cite{he2016deep}  & $256 \times 192$ & 3.9M & 0.82 & 87.7 & 88.2 & 82.6 & 74.7 & 79.3 & 79.8 & 75.4 & 81.6 \\
			 \hline
		\end{tabular}
	}
	\label{tab:posetrack_all}
	\end{center}
	\vspace{-5pt}
\end{table*}

Table~\ref{tab:posetrack_all} compares our proposed ViPNAS with the state-of-the-art methods on  PoseTrack2018~\cite{andriluka2018posetrack} validation set. 

SBL~\cite{xiao2018simple} proposes to add deconvolutional layers to the backbone network, which has been proved effective. We extend ~\cite{xiao2018simple} to include more well-known efficient backbones for comparisons, such as EfficientNet~\cite{tan2019efficientnet}, ShuffleNet~\cite{ma2018shufflenet}, and MobileNet~\cite{howard2019searching}. These models are pre-trained on COCO dataset and fine-tuned on PoseTrack dataset with the same experimental configurations as~\cite{xiao2018simple}. LightTrack~\cite{ning2019lighttrack} is a recently proposed light-weight framework for video pose estimation. The results are obtained using the official codes\footnote{\url{https://github.com/Guanghan/lighttrack}} with the released pre-trained models.

We evaluate our methods on two well-known backbones, \ie ResNet-50~\cite{he2016deep} and MobileNet-V3~\cite{howard2019searching}. For both backbone models, we build the super-network based on the spatial-level search space (Sec.~\ref{sec:spatial_level}) and temporal-level search space (Sec.~\ref{sec:temporal_level}). Please refer to Sec.~\ref{sec:supp_supernetwork} for more details. SBL, LightTrack, and S-ViPNAS directly apply the image-based pose models on each video frame, while T-ViPNAS searches for temporal feature fusion for more efficient pose estimation. \#Param and Flops are calculated by averaging over the whole video frames including both key frames and non-key frames. From Table~\ref{tab:posetrack_all}, we see that ViPNAS achieves the state-of-the-art accuracy with significantly lower model complexity. 
T-ViPNAS significantly boosts the model efficiency and reduces the computation without sacrificing the overall accuracy. For example, T-ViPNAS-MobileNetV3 achieves 10x Flops reduction (0.37 vs 4.1) without accuracy drop (78.2 vs 78.1). 

\begin{figure}[t]
	\centering
	\includegraphics[width=0.4\textwidth]{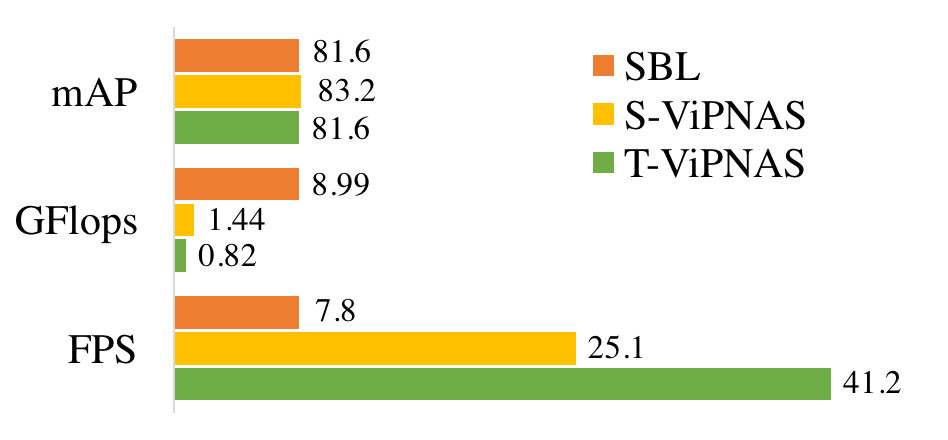}
	\caption{\textbf{Comparisons} among SBL~\cite{xiao2018simple}, S-ViPNAS and T-ViPNAS with ResNet-50 backbone. ViPNAS discovers models with much less computational complexity and significantly higher speed (single core of a 3.2GHz Intel i7-8700 CPU).}
	\label{fig:speed}
\end{figure}

Figure~\ref{fig:speed} compares SBL~\cite{xiao2018simple}, S-ViPNAS and T-ViPNAS with ResNet-50 backbone on PoseTrack2018 validation set. We report mAP, GFlops, and speed (FPS). Speed is evaluated on a single core of an Intel i7-8700 CPU (3.2GHz). We show that T-ViPNAS is significantly faster (41FPS on CPU) than the baseline, with comparable accuracy, making it practical for real-world applications.

\subsection{ViPNAS for image-based pose estimation}
\label{sec:exp-single-frame}

\begin{table*}[tb]
\begin{center}
\caption{\textbf{Comparisons} on COCO2017 dataset. Our approach significantly outperforms other hand-crafted and NAS models in terms of both speed and accuracy on COCO val2017 set and test-dev2017 set. $^\dagger$ means using a stronger person bounding box detector (HTC~\cite{chen2019hybrid}).}
\centering
\resizebox{0.9\linewidth}{!}{
    \begin{tabular}{ c | c | c | c | c | c c c c c c}
    \hline
    ~ & Method & Image Size  & \#Params & GFLOPs & AP & $\text{AP}^{50}$ & $\text{AP}^{75}$ & $\text{AP}^{M}$ & $\text{AP}^{L}$ & $\text{AR}$ \\\hline 
    \multicolumn{11}{c}{COCO Val2017 Set} \\\hline
    \multirow{3}{*}{Hand-Crafted Models} & MobileNet-V3\cite{howard2019searching} & $256 \times 192$ & 5.5M & 4.06  & 64.7 & 86.7  & 72.6 & 61.4  & 70.9  & 76.3 \\ 
    ~ & SBL-50~\cite{xiao2018simple} & $256 \times 192$ & 34.0M & 8.90 & 70.4 & 88.6 & 78.3 & 67.1 & 77.2 & 76.3\\ 
    ~ & HRNet-W32\cite{sun2019deep} & $256 \times 192$ & 28.5M & 7.10 & 74.4 & 90.5 & 81.9 & 70.8 & 81.0 & 79.8\\ \hline
     \multirow{3}{*}{NAS Models} & PoseNFS-3~\cite{yang2019pose} & $384 \times 288$ & 6.1M &  4.0 & 68.0 & -  & -&  - & -  & - \\ 
     ~ & PoseNFS-3~\cite{yang2019pose} & $384 \times 288$ & 15.8M & 14.8  & 73.0 &  - &- &  - &  - & - \\ 
    ~ & AutoPose~\cite{gong2020autopose}$^\dagger$ & $256 \times 192$ & - & 10.65 & 73.6 & 90.6 & 80.1 & 69.8 & 79.7 & 78.1\\   \hline
    \multirow{3}{*}{Ours} & S-ViPNAS-MobileV3 & $256 \times 192$ & \textbf{2.8M} & \textbf{0.69}  & 67.8 & 87.2 & 76.0 & 64.7 & 74.0 & 75.2 \\
    ~ & S-ViPNAS-Res50 & $256 \times 192$ & 13.5M & 1.44 & 71.0 & 89.3 & 78.7 & 67.7 & 77.5 & 76.7  \\ 
    ~ & S-ViPNAS-HRNetW32 & $256 \times 192$ & 16.3M & 5.64  & \textbf{74.7} & 89.9 & 82.0 & 71.0 & 81.5 & 81.2 \\ \hline
    
    \multicolumn{11}{c}{COCO Test-Dev2017 Set} \\\hline
    \multirow{2}{*}{Hand-Crafted Models} & SBL-50~\cite{xiao2018simple} & $256 \times 192$ & 34.0M & 8.90 & 70.0 & 90.9 & 77.9 & 66.8 & 75.8 & 75.6 \\ 
    ~ & HRNet-W32\cite{sun2019deep} & $256 \times 192$ & 28.5M & 7.10 & 73.5 & 91.6 & 81.7 & 70.1 & 79.1 & 80.1 \\ \hline
    \multirow{2}{*}{NAS Models} & PoseNFS-3~\cite{yang2019pose} & $384 \times 288$ & 6.1M & 4.0 & 67.4 & 89.0 & 73.7 & 63.3 & 74.3 & 73.1 \\ 
     ~ & PoseNFS-3~\cite{yang2019pose} & $384 \times 288$ & 15.8M & 14.8  & 72.3 & 90.9 & 79.5 & 68.4 & 79.2 & 77.9 \\ \hline
    \multirow{2}{*}{Ours} & S-ViPNAS-Res50 & $256 \times 192$ & \textbf{13.5M} & \textbf{1.44} & 70.3  & 90.7 & 78.8 & 67.3 & 75.5 & 77.3 \\ 
    ~ & S-ViPNAS-HRNetW32 & $256 \times 192$ & 16.3M & 5.64 & \textbf{73.9} & 91.7 & 82.0 & 70.5 & 79.5 & 80.4 \\ \hline
    \end{tabular}
}
\label{tab:coco_full}
\end{center}
\vspace{-8pt}
\end{table*}

Table~\ref{tab:coco_full} demonstrates the performance of the discovered S-ViPNAS models on COCO2017 dataset, compared with other state-of-the-art hand-crafted methods and concurrent NAS based pose estimators. We report our discovered results based on multiple backbones (\ie HRNet-W32~\cite{sun2019deep}, ResNet-50~\cite{he2016deep} and MobileNetV3~\cite{howard2019searching}). For fair comparisons, we retrain S-ViPNAS models using the same training recipe and use the same Faster-RCNN human detection bounding boxes as SBL~\cite{xiao2018simple} and HRNet~\cite{sun2019deep}.

We see that our discovered S-ViPNAS-HRNetW32 significantly outperforms the popular hand-crafted models and the NAS based models.
Compared with the current state-of-the-art HRNet~\cite{sun2019deep}, we achieve higher accuracy and lower complexity ($5.64$ vs $7.10$ GFlops). Compared with other NAS pose models PoseNFS \cite{yang2019pose} and AutoPose~\cite{gong2020autopose}, ViPNAS also shows superiority in terms of both accuracy and computation complexity.
Note that AutoPose~\cite{gong2020autopose} uses a stronger human detector~\cite{chen2019hybrid} on COCO val2017 set.

We further search for lightweight pose estimators to boost the model efficiency. Based on ResNet-50~\cite{xiao2018simple}, we obtain a 6x smaller ($1.44$ vs $8.90$ GFlops) model (S-ViPNAS-Res50) without sacrificing the accuracy. For MobileNet-V3~\cite{howard2019searching}, our method finds a 5.8x smaller ($0.69$ vs $4.06$ GFlops) model (S-ViPNAS-MobileNet) with 3.1mAP gain ($67.8$ vs $64.7$) on COCO val2017 set.

\begin{figure*}[tb]
\begin{center}
	\includegraphics[width=0.99\textwidth]{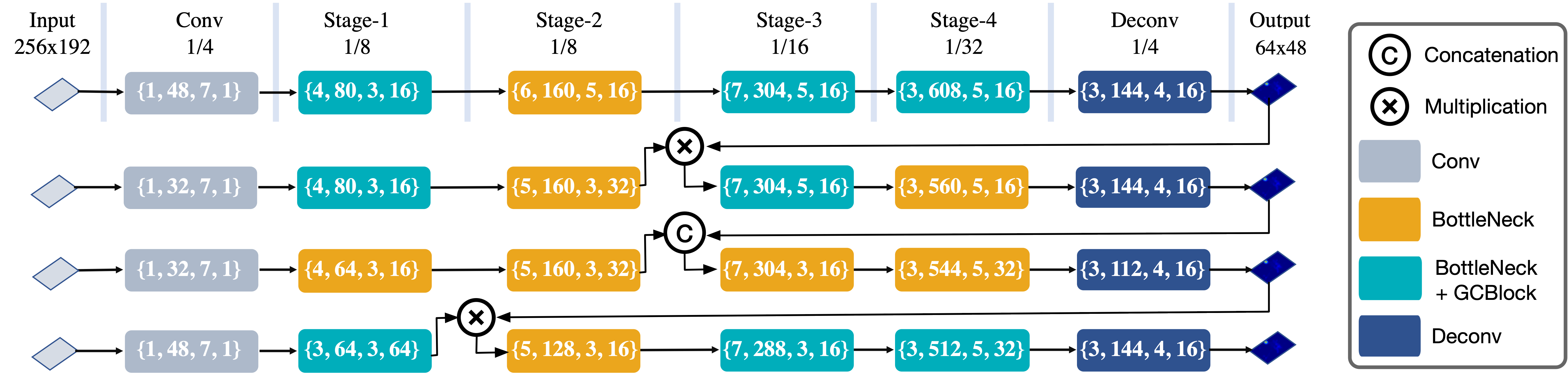}
	\caption{Example of T-ViPNAS with ResNet-50 backbone. \{Depth, Width, Kernel Size, Group\} are listed in the figure. 
	}
	\label{fig:network_architecture}
	\end{center}
	\vspace{-10pt}
\end{figure*}

\subsection{Ablation Study}

\begin{figure*}[tb]
	\centering
	\includegraphics[width=0.9\textwidth]{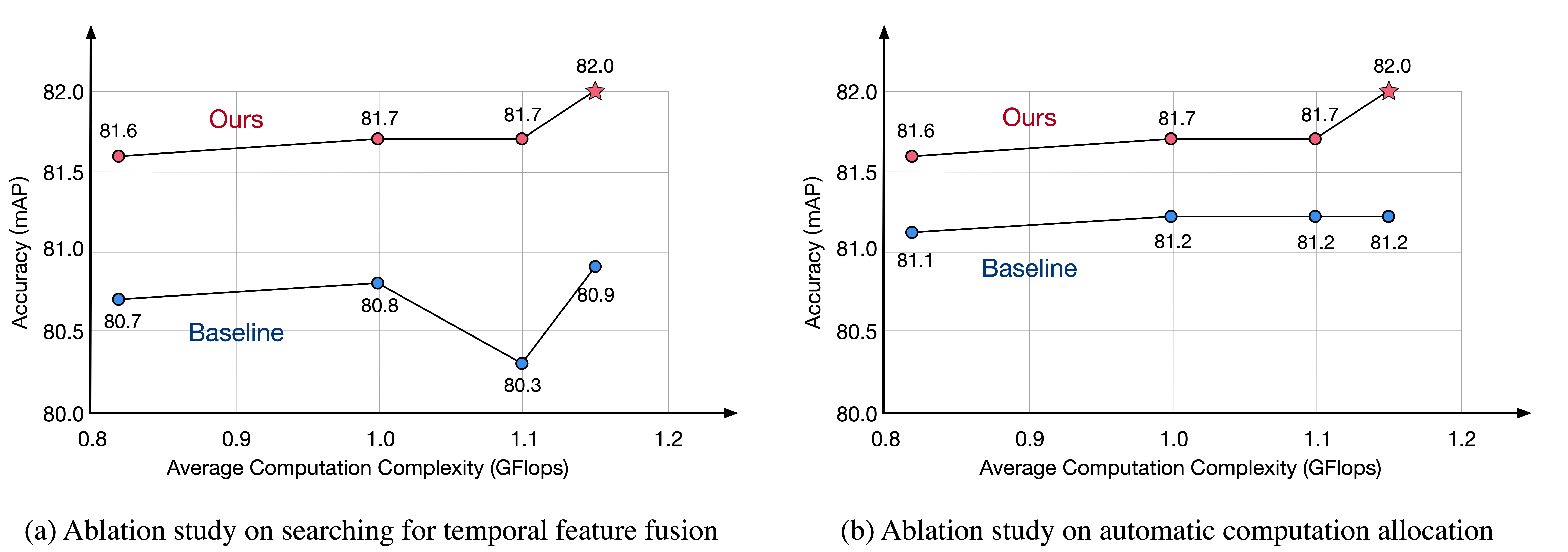}
	\caption{Comparing T-ViPNAS with (a) baselines without temporal feature fusion modules (b) baselines with the same architectures for different frames. We see that our proposed T-ViPNAS consistently improves over the baseline architectures for a range of complexity levels (from 0.8 to 1.2 GFlops). We visualize the architecture of one example T-ViPNAS (red star) in Figure~\ref{fig:network_architecture}. }
	\label{fig:ablation}
\end{figure*}

\textbf{Effect of temporal-level search}. To validate the effect of temporal-level search, we search S-ViPNAS under the constraints of the same model complexity as T-ViPNAS. We apply the image-based S-ViPNAS models independently for each frame. As shown in Table~\ref{tab:samllS}, we see that given the same Flops constraints, T-ViPNAS discovers better model architectures with higher accuracy (81.6 vs 80.3 mAP for ResNet-50 based models (-a) and 78.2 vs 77.2 mAP for MobileNet-V3 based models (-b)).

\begin{table}[tb]
	\begin{center}
	\caption{Effect of temporal-level NAS for video pose estimation on PoseTrack2018 dataset. Given the same GFlops constraints, T-ViPNAS discovers better architectures with higher accuracy.}
		\begin{tabular}{c|c|c|c}
			\hline
			Method & Backbone &  GFLOPs  &  mAP  \\\hline
			S-ViPNAS-a & ResNet-50  & 0.82  & 80.3 \\
            T-ViPNAS-a & ResNet-50   & 0.82  & \textbf{81.6} \\ \hline
		    S-ViPNAS-b & MobileNet-V3   & 0.37  & 77.2 \\
		    T-ViPNAS-b & MobileNet-V3  & 0.37  & \textbf{78.2} \\ \hline
		\end{tabular}
	\label{tab:samllS}
	\end{center}
	\vspace{-15pt}
\end{table}

\textbf{Effect of temporal feature fusion.}
As shown in Figure~\ref{fig:ablation}(a), we explore the effect of temporal feature fusion on the PoseTrack2018 validation set. We search for four groups of T-ViPNAS models with ResNet-50 backbone in a range of average computation complexity levels (from 0.8 to 1.2 GFLOPs) for comparisons. The number of propagation frames is set as $T=3$, so for each group, we have 4 different models (\ie 1 S-ViPNet and 3 T-ViPNets) in total.

To validate the effectiveness of temporal feature fusion, we remove the temporal feature fusion from T-ViPNAS (red) in each group, keep the model architecture the same, and re-train them based on single images (blue). We see that our T-ViPNAS consistently improves over the baselines for various Flops requirements. Our experiments show that temporal fusion captures the consistency among adjacent frames and propagates poses efficiently using extremely lightweight models.

\textbf{Effect of automatic computation allocation.} 
As shown in Figure~\ref{fig:ablation}(b), we further explore the effect of automatic computation allocation on the PoseTrack2018 validation set. For comparisons, we search for the temporal models sharing both the spatial and temporal architectures (blue) under the same Flops constraints as our T-ViPNAS (red). We find that our T-ViPNAS consistently improves over the baseline architectures by at least 0.5\% mAP, demonstrating the effectiveness of automatic computation allocation that searches for frame-specialized models. Example architectures of our discovered models are visualized in Figure~\ref{fig:network_architecture}. 

\textbf{Effect of the number of propagation frames.} We evaluate the transferability of our proposed ViPNAS training scheme to various propagation frames $T$. During training of T-ViPNAS (Sec.~\ref{sec:train_tvip}) with ResNet-50 backbone, we set the number of non-key frames as $T=3$, but search on different propagation lengths without re-training the super-network. As shown in Table~\ref{tab:propagation_number}, we set the constraints of the average model computation complexity to be 1.0 GFlops, and search for different propagation frame numbers, \ie $T=2$, $T=3$ or $T=4$ frames. We see that our ViPNAS is relatively robust to the number of propagation frames.

\begin{table}[tb]
	\begin{center}
	\caption{Effect of the number of propagation frames. We experiment with training the super-network for $T=3$ frames and searching for $T=\{2,3,4\}$ frames with 1.0 GFlops complexity constraint on average.}
		\begin{tabular}{c|c|c}
			\hline
			\#Propagation Frames ($T$) & GFLOPs & mAP   \\\hline
			2  & 1.0 & 81.7 \\ 
			3  & 1.0 & 81.7 \\ 
			4  & 1.0 & 81.4 \\ \hline
		\end{tabular}
	\label{tab:propagation_number}
	\end{center}
	\vspace{-20pt}
\end{table}

\section{Conclusion}

In this paper, we propose ViPNAS for online video pose estimation, trading-off between accuracy and the computation cost. ViPNAS automatically allocates computation resources (\ie Flops) for different frames to achieve the overall optimum. By designing the novel spatial-temporal search space, we can simultaneously search for CNN architectures and temporal connections, \ie the fusion operations and the feature fusion sites. Empirical experiments demonstrate that our proposed ViPNAS successfully discovers the architecture that achieves the state-of-the-art accuracy with CPU real-time performance.

\textbf{Acknowledgement.} This work is supported in part by the General Research Fund through the Research Grants Council of Hong Kong under Grants (Nos. 14202217, 14203118, 14208619), in part by Research Impact Fund Grant No. R5001-18. Ping Luo is supported by the Research Donation from SenseTime and the General Research Fund of HK No.27208720. Wanli Ouyang is supported by the Australian Research Council Grant DP200103223 and Australian Medical Research Future Fund MRFAI000085.

{\small
\bibliographystyle{ieee_fullname}
\bibliography{egbib}
}

\appendix
\section*{\Large Appendix}
\setcounter{table}{0}
\renewcommand{\thetable}{A\arabic{table}}
\setcounter{figure}{0}
\renewcommand{\thefigure}{A\arabic{figure}}

\section{Implementation Details}
\label{sec:supp_implementation}
\subsection{Spatial Search Space}
\label{sec:supp_spatial_search_space}
We give implementation details of the spatial search space of ViPNAS in this section.

\begin{figure}[tb]
	\centering
	\includegraphics[width=0.49\textwidth]{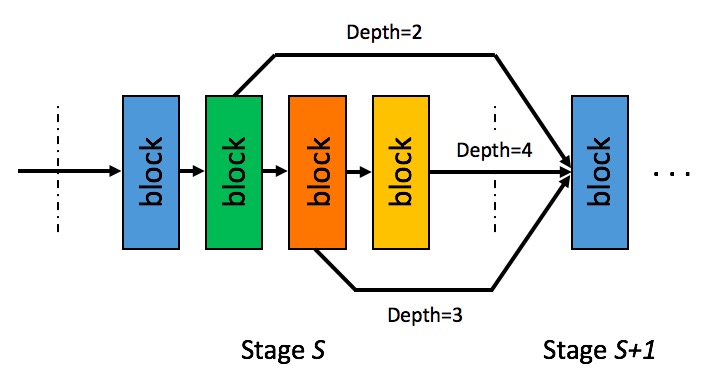}
	\caption{\textbf{Elastic Depth.} The first $D$ blocks are activated if the depth $D$ is selected in stage $S$.}
	\label{fig:depth}
	\vspace{10pt}
\end{figure}

\textbf{Elastic Depth.} 
Elastic depth allows dynamic numbers of blocks in each stage. For example, the maximum number of the blacks in stage $S$ is 4 as shown in Figure~\ref{fig:depth}. When the depth $D$ ($D\leq4$) is selected, the first $D$ blocks are activated and the rest ($4-D$) blocks are skipped. Note that the minimum depth of any stage should be no less than 1 ($D\geq1$), as the first block may change the spatial resolution of the feature maps.

\begin{figure*}[tb]
	\centering
	\includegraphics[width=0.99\textwidth]{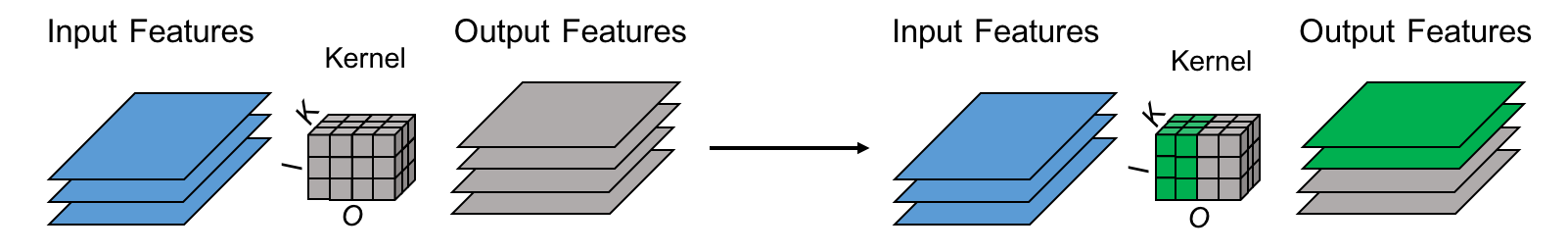}
	\caption{\textbf{Elastic Width.} Given the input channels $I$ and kernel size $K \times K$, the first $W$ output channels out of $O$ is kept if the width W is selected. The filter is tailored from the shape $O \times I \times K \times K$ to $W \times I \times K \times K$.}
	\label{fig:width}
	\vspace{10pt}
\end{figure*}

\textbf{Elastic Width.}
Elastic width allows dynamic numbers of output channels in each block. For a convolutional layer, the shape of the filter is $O \times I \times K \times K$ given the input channels $I$, output channels $O$, and kernel size $K \times K$. When the output channel $W$ ($W \leq O$) is selected, the filter is tailored to the shape of $W \times I \times K \times K$ as shown in Figure~\ref{fig:width}. We keep the first $W$ out of $O$ in the dimension of output channels.

\begin{figure}[tb]
	\centering
	\includegraphics[width=0.35\textwidth]{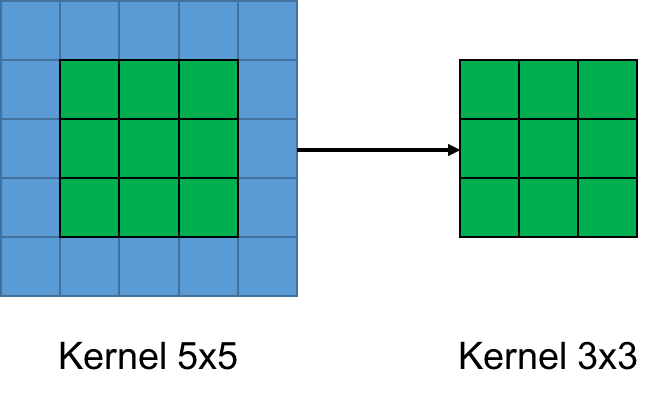}
	\caption{\textbf{Elastic Kernel Size.} The centering $K \times K$ kernel is reserved if the kernel size $K$ is selected.}
	\label{fig:ks}
	\vspace{10pt}
\end{figure}

\textbf{Elastic Kernel Size.}
Elastic kernel size allows dynamic kernel sizes of convolutional layers in each block. The weights of the kernels are shared. As shown in Figure~\ref{fig:ks}, we directly extract a $K \times K$ kernel filter from the centering of the super-network kernel filter, when the kernel size $K$ is selected. This enables the weight sharing for kernels of different sub-networks, which has been shown simple but effective in our experiments. To avoid imbalance and biases of kernel extraction, we set the stride of the kernel size choice as 2, keeping all the selected kernels center-aligned.

\begin{figure*}[h]
	\centering
	\includegraphics[width=0.99\textwidth]{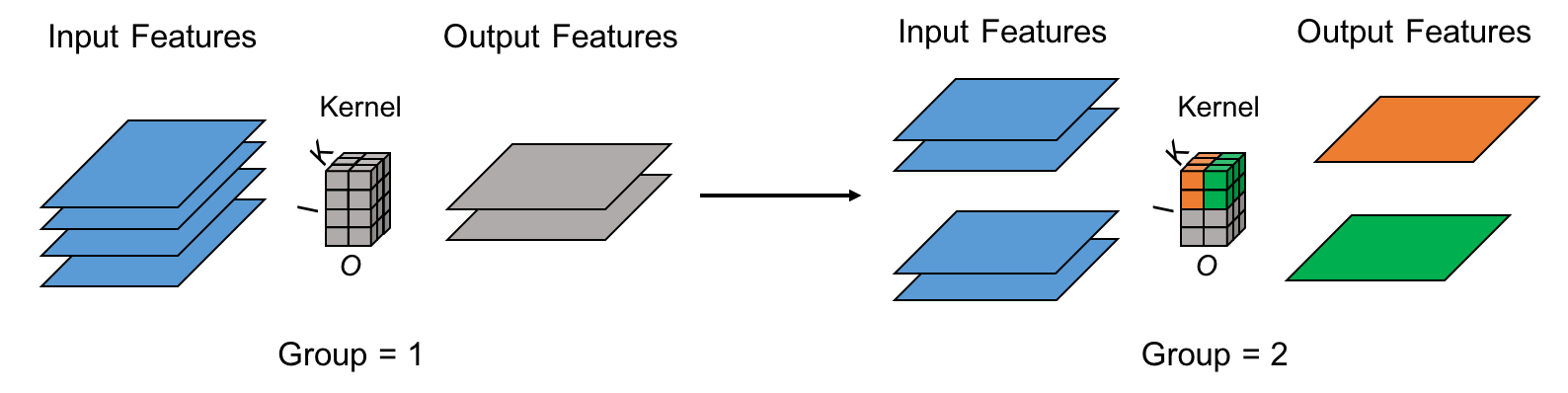}
	\caption{\textbf{Elastic Group Number.} An example of Group=2 is illustrated in the figure. Given the input channels $I$, output channels $O$, and kernel size $K \times K$, the filter is tailored from the shape $O \times I \times K \times K$ to $O \times \frac{I}{2} \times K \times K$. Two groups of filters with shape $\frac{O}{2} \times \frac{I}{2} \times K \times K$ are applied and are concatenated in the dimension of output channels.}
	\label{fig:group}
	\vspace{10pt}
\end{figure*}

\textbf{Elastic Group Number.}
Elastic group number allows dynamic group numbers of convolutional layers in each block. A convolutional layer has a filter with the shape $O \times I \times K \times K$ given the input channels $I$, output channels $O$ and kernel size $K \times K$. For example, when the group number is 2 (as shown in Figure~\ref{fig:group}), two filters with shape $\frac{O}{2} \times \frac{I}{2} \times K \times K$ are applied. In the figure, we concatenate the two groups of filters in the dimension of output channels for better illustration. We tailor the original filter to the shape of $O \times \frac{I}{2} \times K \times K$ and keep the first half in the dimension of input channels.

\begin{figure}[tb]
	\centering
	\includegraphics[width=0.49\textwidth]{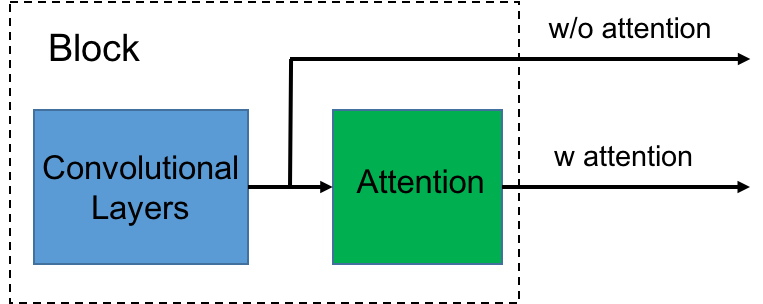}
	\caption{\textbf{Elastic Attention Module.} The attention module is applied if attention is selected, and is skipped if not.}
	\label{fig:attention}
	\vspace{10pt}
\end{figure}

\textbf{Elastic Attention Module.}
Elastic attention module allows the network to choose whether or not to use the attention module in each block. As shown in Figure~\ref{fig:attention}, the attention module is used if attention module is selected. We skip the attention module and identity mapping is applied if attention module is not selected. The attention module will keep both the spatial resolution and the feature channels the same before and after.

\subsection{Super-Network Design}
\label{sec:supp_supernetwork}
In this section, we introduce the structure of our super-network as well as the concrete search space designs for each super-network. As the search space increases with the exponential explosion, directly searching for block-level network architecture is hard. In our experiments, we explicitly enforce the same width, kernel size, group number and attention module for all the blocks in the same stage and search for stage-wise optimum.

\textbf{MobileNet-V3~\cite{howard2019searching}.}
Our MobileNet-V3 based super-network consists of one convolutional layer, six stages, and three deconvolutional layers (followed by one $1\times1$ convolutional layer for output). Each stage contains a stack with mobile blocks~\cite{howard2019searching}, which consists of one $1\times1$ expansion convolution, a middle convolution and one $1\times1$ projection convolution. We search for the kernel size and the group number of the middle convolution in mobile blocks. The expansion convolution expands the input features to a higher-dimensional feature space. We search the expansion ratio, which is similar to the elastic width. The detailed search space is summarized in Table~\ref{tab:mbv3}.

\textbf{ResNet-50~\cite{he2016deep}.}
Following SBL~\cite{xiao2018simple}, our ResNet-50 based super-network consists of one convolutional layer, four stages and three deconvolutional layers. Each block in stages is Bottleneck~\cite{he2016deep}, which contains one $1\times1$ convolution followed by a middle convolution and another $1\times1$ convolution. Similar to our MobileNet-V3 based super-network design, we search for the kernel size and the group number of the middle convolution in the Bottleneck. We also search for whether to use a GC attention module~\cite{cao2019gcnet} in each block. Table~\ref{tab:res50} specifies the search space of our ResNet-50 based super-network.

\textbf{HRNet-W32\cite{sun2019deep}.}
We conduct experiments based on HRNet-W32 to further demonstrate the effectiveness of our proposed ViPNAS. Our HRNet-W32 based super-network consists of two convolutional layers followed by several Bottleneck blocks, three multi-resolution stages, and one $1\times1$ convolutional head for output. Each multi-resolution stage contains parallel branches with different spatial resolution, and each branch includes several BasicBlock~\cite{he2016deep}. Both the convolutions in BasicBlock apply the same width, kernel size, and group number. We search the configurations of each stage and each branch for the best performance. Table~\ref{tab:hrw32} displays the detailed search space of our HRNet-W32 based super-network.

Our search space is discrete. Take ResNet-50 backbone as an example, we set the search step to be 1 for depth, 16 for width, 2 for kernel size and 16 for group.

\begin{table*}[tb]
	\begin{center}
	\caption{\textbf{MobileNet-V3~\cite{howard2019searching}} based search space. [min, max] indicates the range of each search space. Expansion ratio indicates the feature channel expansion rate in the middle of mobile blocks, and resolution indicates the ratio between the shapes of current features and those of input images. Kernel size and group number of the middle convolution in mobile blocks are searched.}
	 \scalebox{0.95}{
		\begin{tabular}{c|c|c|c|c|c|c|c|c}
			\hline
    			Stage & Operator & Depth & Width & Kernel Size & Group & Attention (SE~\cite{hu2018squeeze}) & Expansion Ratio & Resolution \\  \hline
		      & Conv & - & [16, 16] & [3, 3] & [1, 1] & - & - & 1/2 \\  \hline
		    1 & Mobile Block & [1, 1] & [16, 16] & [3, 3] & [2, 16] & [0, 1] & [1, 1] & 1/2 \\
		    2 & Mobile Block & [2, 4] & [24, 24] & [3, 7] & [9, 144] & [0, 1] & [3, 6] & 1/4 \\
		    3 & Mobile Block & [2, 4] & [40, 40] & [3, 7] & [15, 240] & [0, 1] & [3, 6] & 1/8 \\
		    4 & Mobile Block & [2, 4] & [80, 80] & [3, 7] & [30, 480] & [0, 1] & [3, 6] & 1/16  \\
		    5 & Mobile Block & [2, 4] & [112, 112] & [3, 7] & [42, 672] & [0, 1] & [3, 6] & 1/16  \\
		    6 & Mobile Block & [2, 4] & [160, 160] & [3, 7] & [60, 960] & [0, 1] & [3, 6] & 1/32  \\  \hline
		      & Deconv & - & [256, 256] & [4, 4] & [32, 256] & - & - & 1/4 \\  \hline
			
		\end{tabular}
	}
	\label{tab:mbv3}
	\end{center}
\end{table*}

\begin{table*}[tb]
	\begin{center}
	\caption{\textbf{ResNet-50~\cite{he2016deep}} based search space. [min, max] indicates the range of each search space, and expansion ratio indicates the feature channel expansion rate in the middle of Bottleneck. The first convolution and max pooling with stride 2 down-sample the spatial resolution to 1/4 of the input image. Kernel size and group number of the middle convolution in Bottleneck are searched.}
	 \scalebox{0.99}{
		\begin{tabular}{c|c|c|c|c|c|c|c|c}
			\hline
    			Stage & Operator & Depth & Width & Kernel Size & Group & Attention (GC~\cite{cao2019gcnet}) & Expansion Ratio & Resolution \\  \hline
		      & Conv+Pool & - & [32, 64] & [7, 7] & [1, 1] & - & - & 1/4 \\  \hline
		    1 & Bottleneck & [3, 4] & [64, 80] & [3, 5] & [16, 64] & [0, 1] & [1, 1] & 1/8 \\
		    2 & Bottleneck & [4, 6] & [128, 160] & [3, 5] & [16, 64] & [0, 1] & [1, 1] & 1/8 \\
		    3 & Bottleneck & [6, 8] & [256, 320] & [3, 5] & [16, 64] & [0, 1] & [1, 1] & 1/16 \\
		    4 & Bottleneck & [3, 4] & [512, 640] & [3, 5] & [16, 64] & [0, 1] & [1, 1] & 1/32  \\  \hline
		      & Deconv & - & [64, 256] & [4, 4] & [16, 64] & - & - & 1/4 \\  \hline
			
		\end{tabular}
	}
	\label{tab:res50}
	\end{center}
\end{table*}

\begin{table*}[tb]
	\begin{center}
	\caption{\textbf{HRNet-W32~\cite{sun2019deep}} based search space. HRNet includes parallel branches with different resolution in stages, which indicates the ratio between the spatial shape of current features and input images. We search depth of each stage, and search width and attention of each branch. Kernel size and group number of the middle convolution in Bottleneck and both the convolutions in BasicBlock are searched.}
	\vspace{12pt}
	 \scalebox{0.99}{
		\begin{tabular}{c|c|c|c|c|c|c|c|c}
			\hline
    			Stage & Depth & Branch & Operator & Width & Kernel Size & Group & Attention (SE~\cite{hu2018squeeze}) & Resolution \\  \hline
    	     & - &  & Conv & [16, 64] & [3, 3] & [1, 1] & - & 1/4 \\  \hline
		    1 & [2, 4] & 1 & Bottleneck & [16, 64] & [3, 3] & [1, 16] & [0, 1] & 1/4 \\  \hline
		    \multirow{2}{*}{2} & \multirow{2}{*}{[4, 4]} & 1 & BasicBlock & [8, 32] & [3, 3] & [1, 32] & [0, 1] & 1/4 \\  \cline{3-9}
		    ~ & ~ & 2 & BasicBlock & [16, 64] & [3, 3] & [1, 64] & [0, 1] & 1/8  \\  \hline
		    \multirow{3}{*}{3} & \multirow{3}{*}{[8, 16]} & 1 & BasicBlock & [8, 32] & [3, 3] & [1, 32] & [0, 1] & 1/4 \\\cline{3-9}
		    ~ & ~ & 2 & BasicBlock & [16, 64] & [3, 3] & [1, 64] & [0, 1] & 1/8 \\\cline{3-9}
		    ~ & ~ & 3 & BasicBlock & [32, 128] & [3, 3] & [1, 128] & [0, 1] & 1/16 \\  \hline
		    \multirow{4}{*}{4} & \multirow{4}{*}{[8, 12]} & 1 & BasicBlock & [8, 32] & [3, 3] & [1, 32] & [0, 1] & 1/4 \\\cline{3-9}
		    ~ & ~ & 2 & BasicBlock & [16, 64] & [3, 3] & [1, 64] & [0, 1] & 1/8 \\\cline{3-9}
		    ~ & ~ & 3 & BasicBlock & [32, 128] & [3, 3] & [1, 128] & [0, 1] & 1/16 \\\cline{3-9}
		    ~ & ~ & 4 & BasicBlock & [64, 256] & [3, 3] & [1, 256] & [0, 1] & 1/32 \\  \hline
		\end{tabular}
	}
	\label{tab:hrw32}
	\end{center}
\end{table*}

\section{Qualitative Results}
\begin{figure*}[t]
	\centering
	\includegraphics[width=0.99\textwidth]{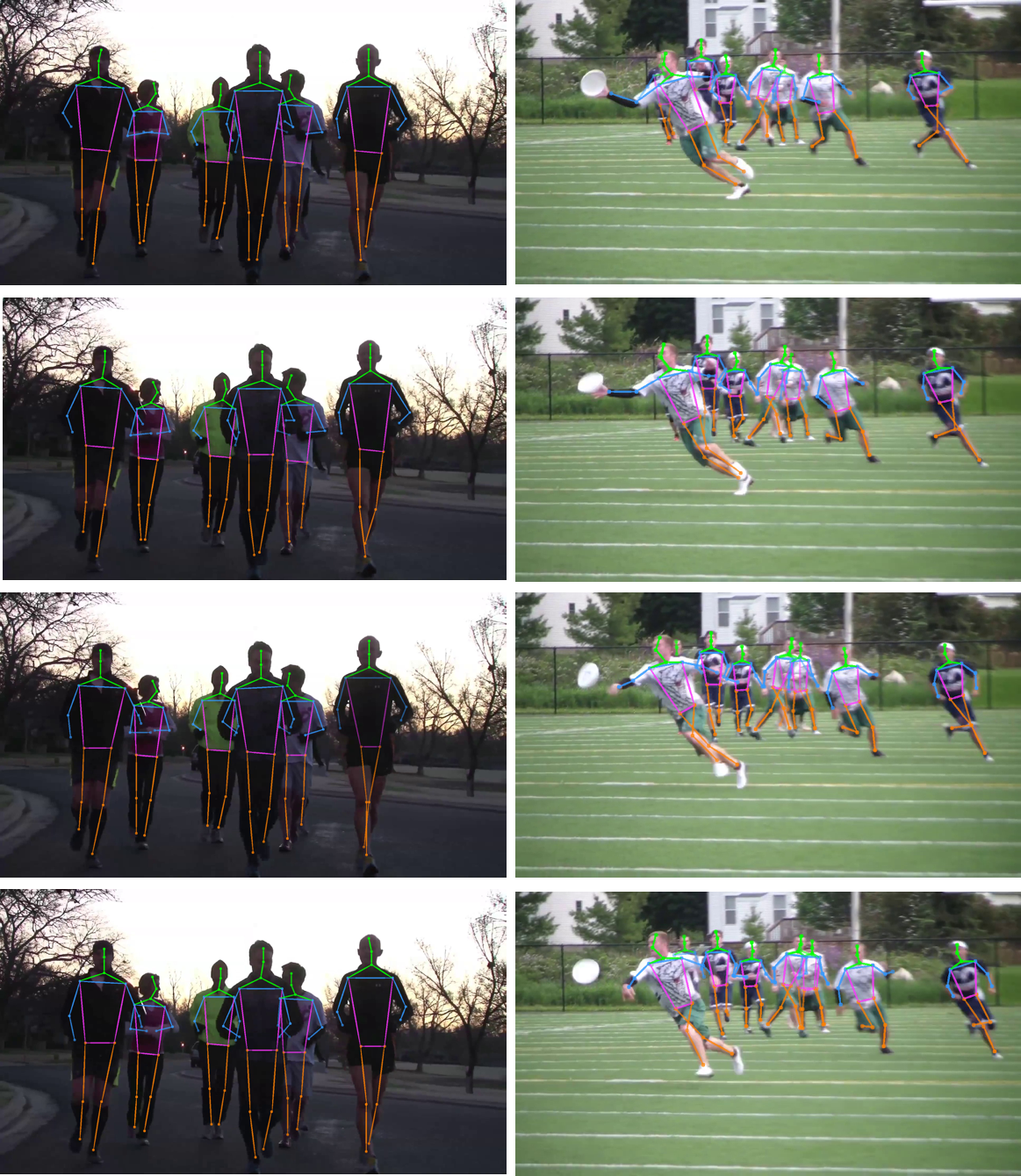}
	\caption{Qualitative results of T-ViPNAS-Res50 on four adjacent frames. S-ViPNet localizes human poses on the first frame, and three different T-ViPNets propagate poses on the following frames. Our proposed ViPNAS is robust to occlusion, motion blur and unusual illumination, and achieves state-of-art accuracy with CPU real-time performance.}
	\label{fig:vis}
\end{figure*}

Figure~\ref{fig:vis} shows the qualitative results of our T-ViPNAS-Res50 on four adjacent frames. S-ViPNet localizes human poses on the first frame (key frame), and three different T-ViPNets propagate poses on the following frames (non-key frame). Our lightweight models keep the temporal consistency and are robust to occlusion, motion blur and unusual illumination. ViPNAS achieves state-of-the-art accuracy with CPU real-time performance.

\begin{figure*}[t]
	\centering
	\includegraphics[width=0.99\textwidth]{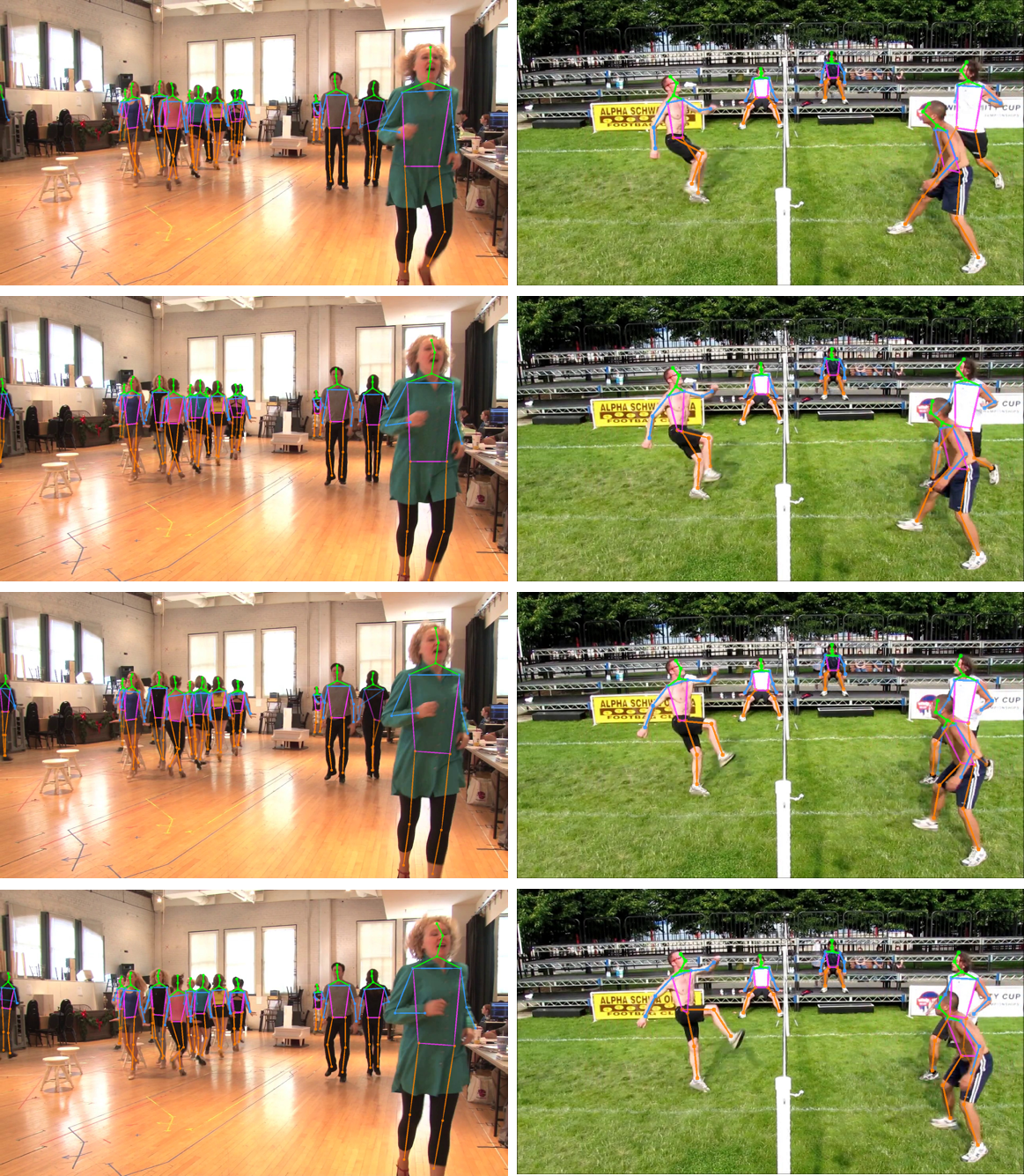}
\end{figure*}

\end{document}